\documentclass[10pt,twocolumn,letterpaper]{article}
\pdfoutput=1
\usepackage{iccv} 
\usepackage{booktabs}
\usepackage{colortbl}
\usepackage{graphicx}
\usepackage{multirow}
\usepackage{xcolor}
\usepackage{tabularx}
\usepackage{array}
\usepackage{pifont} 
\usepackage{hyperref}

\definecolor{iccvblue}{rgb}{0.21,0.49,0.74}

\def\confName{ICCV}
\def\confYear{2025}
\newcommand{\xmark}{\ding{55}}%

\newcommand{\zql}[1]{\textcolor{black}{#1}}
\newcommand{\wq}[1]{\textcolor{black}{#1}}
\title{H²VU-Benchmark: A Comprehensive Benchmark for Hierarchical Holistic Video Understanding}

\author{
    Qi Wu$^*$ \hspace{0.5em}
    Quanlong Zheng$^*$ \hspace{0.5em}
    Yanhao Zhang$^\dagger$$^\spadesuit$\hspace{0.5em}
    Junlin Xie \hspace{0.5em}
    Jinguo Luo \hspace{0.5em} \\
    Kuo Wang \hspace{0.5em}
    Peng Liu \hspace{0.5em}
    Qingsong Xie \hspace{0.5em}
    Ru Zhen\hspace{0.5em}
    Haonan Lu$^\dagger$\hspace{0.5em}
    Zhenyu Yang \hspace{0.5em} \\
    OPPO AI Center \\
    \{wuqi,zhengquanlong,zhangyanhao\}@oppo.com \\
    \footnotesize{
    $^{*}$~Equal Contribution \;
    $^{\spadesuit}$~Project Leader \;
    $^{\dagger}$~Corresponding Author \;
    }
}

\begin{document}
\maketitle
\begin{figure*}[t]
  \centering
\includegraphics[width=\linewidth]{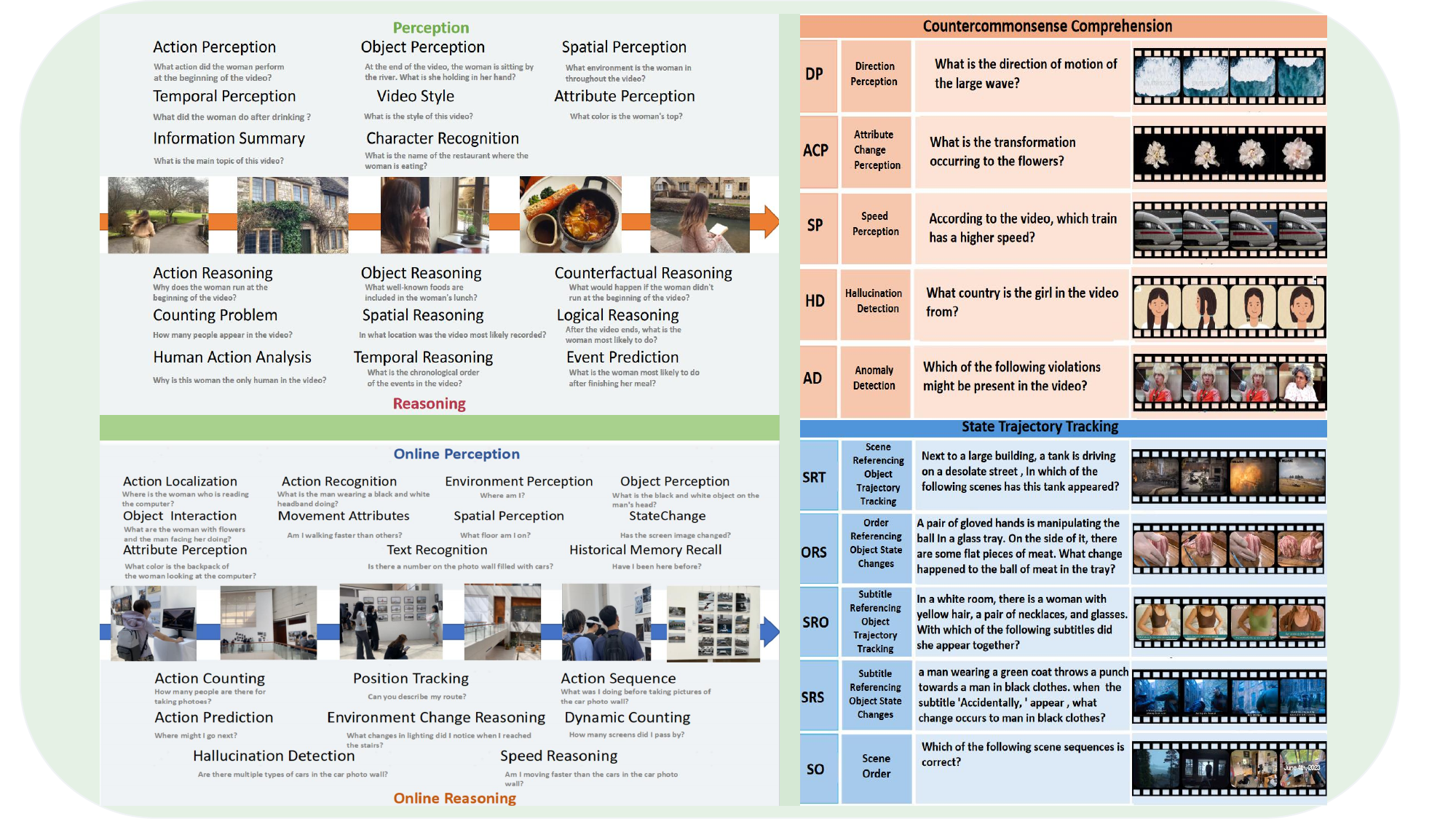}

   \caption{Examples of each task in H²VU-Bench. Based on the video input and text prompt, Multi-Modal Language Models (MLLMs) are required to select the correct option.}
   \label{fig:onecol}
\end{figure*}
\begin{abstract}

With the rapid development of multimodal models, the demand for assessing video understanding capabilities has been steadily increasing. However, existing benchmarks for evaluating video understanding exhibit significant limitations in coverage, task diversity, and scene adaptability. These shortcomings hinder the accurate assessment of models’ comprehensive video understanding capabilities. To tackle this challenge, we propose a hierarchical and holistic video understanding (H²VU) benchmark designed to evaluate both general video and online streaming video comprehension.This benchmark contribute three key features:
(1) Extended video duration: Spanning videos from brief 3-second clips to comprehensive 1.5-hour recordings, thereby bridging the temporal gaps found in current benchmarks.
(2) Comprehensive assessment tasks: beyond traditional perceptual and reasoning tasks, we have introduced modules for countercommonsense comprehension and trajectory state tracking. These additions test the models' deep understanding capabilities beyond mere prior knowledge.
(3) Enriched video data: To keep pace with the rapid evolution of current AI agents, we have expanded first-person streaming video datasets. This expansion allows for the exploration of multimodal models' performance in understanding streaming videos from a first-person perspective. 
Extensive results from H²VU reveal that existing MLLMs possess substantial potential for improvement in our newly proposed evaluation tasks. We expect that H²VU will facilitate advancements in video understanding research by offering a comprehensive and in-depth analysis of MLLMs.More details can be found on the paper's GitHub page:https://github.com/siriusrecco/H2VU-BenchMark

\end{abstract}    
\section{Introduction}
\label{sec:intro}

Video content, as a powerful multimodal information medium in the digital era, has deeply permeated core aspects of knowledge dissemination, experience sharing, and entertainment consumption. With global internet video traffic projected to exceed 4 ZB by 2027 \cite{chen2024comprehensive, istrate2024environmental}, this exponential growth in video data is reshaping communication, learning, and connectivity in the digital age. In response to the increasing demand for analyzing vast amounts of video content, \zql{multimodal large language models (MLLMs) have shown notable success in the field of video understanding \cite{Ormazabal2024RekaCF,OpenAI2023ChatGPT,team2023gemini,Alayrac2022FlamingoAV,zhang2025videollama,chen2024internvl2,bai2025qwen2,wang2024qwen2,li2024videochat,zhang2024video,damonlpsg2023videollama}.}

\zql{Various video benchmarks\cite{zhang2024mathverse,Yu2023MMVetEL,mathvista,fu2023mme,jang2017tgif,TextVQA,li2020hero,wu2021star_situated_reasoning,guan2023hallusionbench} have been proposed to accurately evaluate the capabilities of videoLLMs. However, existing dominant video benchmarks possess several limitations\cite{Liu2024TempCompassDV,li2023vitatecs,li2023mvbench,lin2024streamingbench,fu2024video,ning2023video,li2025ovo,li2024videovista}.}  
\zql{Typically, these benchmarks focus on short videos ranging from a few seconds to several minutes\cite{li2024videovista,lin2024streamingbench}, with some extending to 10 minutes or even 1.5 hours\cite{fu2023mme,wang2024lvbench,zhou2024mlvu,song2024moviecha,chai2024auroracap}. Despite this, they still fall short of adequately representing real-world applications.}
Second\zql{ly},current benchmarks are deficient in delving into the temporal content of videos. To some degree, they rely on the prior knowledge intrinsically embedded within MLLMs for question-answering purposes. \zql{They inadequately evaluate complex video understanding and dynamic state tracking capabilities. Furthermore, most benchmarks are limited to third-person videos, neglecting the unique challenges posed by first-person streaming video scenarios}
, resulting in a significant gap between their demonstrated capabilities and the requirements of real-world assistants or autonomous agents.

To address these issues, we propose a new video understanding benchmark aimed at comprehensively evaluating \zql{MLLMs' real-world video understanding ability.}
Our primary contributions are as follows.
Extended Video \zql{Duration}: 
\zql{O}ur benchmark encompasses a diverse range from a few seconds to 1.5 hours, significantly expanding the temporal scope. This extension \zql{evaluates} models' ability to capture short-term dynamics and their capacity to model long-term dependencies.
\zql{Advanced Task Complexity:}
Building on traditional perceptual and reasoning tasks, we introduce two new modules.
Counterfactual Reasoning: \zql{This module assesses models' vision-oriented understanding ability through counterfactual tasks that defy common sense, such as implausible causal relationships or events that contravene physical laws.} 
Trajectory State Tracking: This \zql{module} evaluates models' abilities to track and predict the states and trajectories of targets in complex dynamic scenes.
\zql{Diversified Real-world Data:}
As AI agents increasingly function as real-world assistants or autonomous agents\cite{yang2024v,huang2022planner}, we incorporate first-person streaming video data. Videos from a first-person perspective contain rich interactive information and dynamic scenes, better simulating real-world streaming data processing needs. Through this expansion, we further explore multimodal models' performance in understanding first-person streaming video. \zql{The sample demonstrations in H²VU-Bench are illustrated in fig \ref{fig:onecol}.}

Our benchmark includes various task settings, covering diverse video scenarios and perspectives, forming a comprehensive evaluation framework. We utilize this benchmark to evaluate a range of state-of-the-art MLLMs, including GPT-4o and Gemini-1.5 Pro and Flash, as well as open-source video models. Experimental results indicate significant differences between Countercommonsense Comprehension tasks and state trajectory tracking tasks compared to common tasks. This disparity not only highlights the current models' deficiencies in overcoming the constraints of prior knowledge to respond based on video content but also underscores the inadequacies in their temporal understanding and objective sustained perception capability. Moreover, some models that perform well on general video tasks exhibit poor performance in streaming first-person perspective video tasks, indicating that video understanding models still face challenges in real-world applications.


\section{Related Works}
\subsection{Video Large Language Models}

\zql{As the multimodal large languae models' understanding capabilities become increasingly advanced, they also exhibit exceptional understanding abilities in video comprehension by extracting certain video frames as images for input. For example, VideoChat-Flash\cite{li2024videochat}, Video-LLaMA\cite{touvron2023llama}, and Video-ChatGPT\cite{maaz2024videogpt+} utilize CLIP-ViT\cite{radford2021learning} embeddings of selected video frames, projecting them into the LLM embedding space through amultilayer perceptrons (MLPs). These embeddings are then concatenated with text embeddings to enhance video understanding.}



\zql{However, the context-length limitations of multimodal large language models (MLLMs) constrain their capability to effectively understand long videos that require more frames and extended contextual comprehension.} 
Video-LLaMA 3\cite{zhang2025videollama} enhances spatiotemporal modeling and audio comprehension through the incorporation of a customized spatiotemporal convolution connector (STC) and an audio branch. Qwen2.5-VL\cite{bai2025qwen2} improves video understanding by employing dynamic frame rate sampling and updating the mRoPE in the temporal dimension, enabling the model to learn multi-resolution video with temporal sequences and velocity. 
\zql{InternVL2.5\cite{chen2024internvl2}, which is built upon the InternVL 2.0\cite{chen2024internvl} architecture, delivers performance on par with commercial models. This is achieved through improvements in training strategies, refinements in testing protocols, and enhancements in data quality.}
LLaVA-Video\cite{zhang2024video} utilizes a high-quality synthetic dataset specifically created for video instruction-following tasks, achieving performance that matches or surpasses certain open-source models. Despite these advancements, the potential of multimodal large language models in processing sequential visual data remains underexplored. Consequently, we introduce the H²VU-Benchmark for \zql{a comprehensive evaluation of both commercial and open-source MLLMs for video understanding.}
\begin{figure*}
  \centering
  \begin{subfigure}{0.58\linewidth}
    \includegraphics[width=\linewidth]{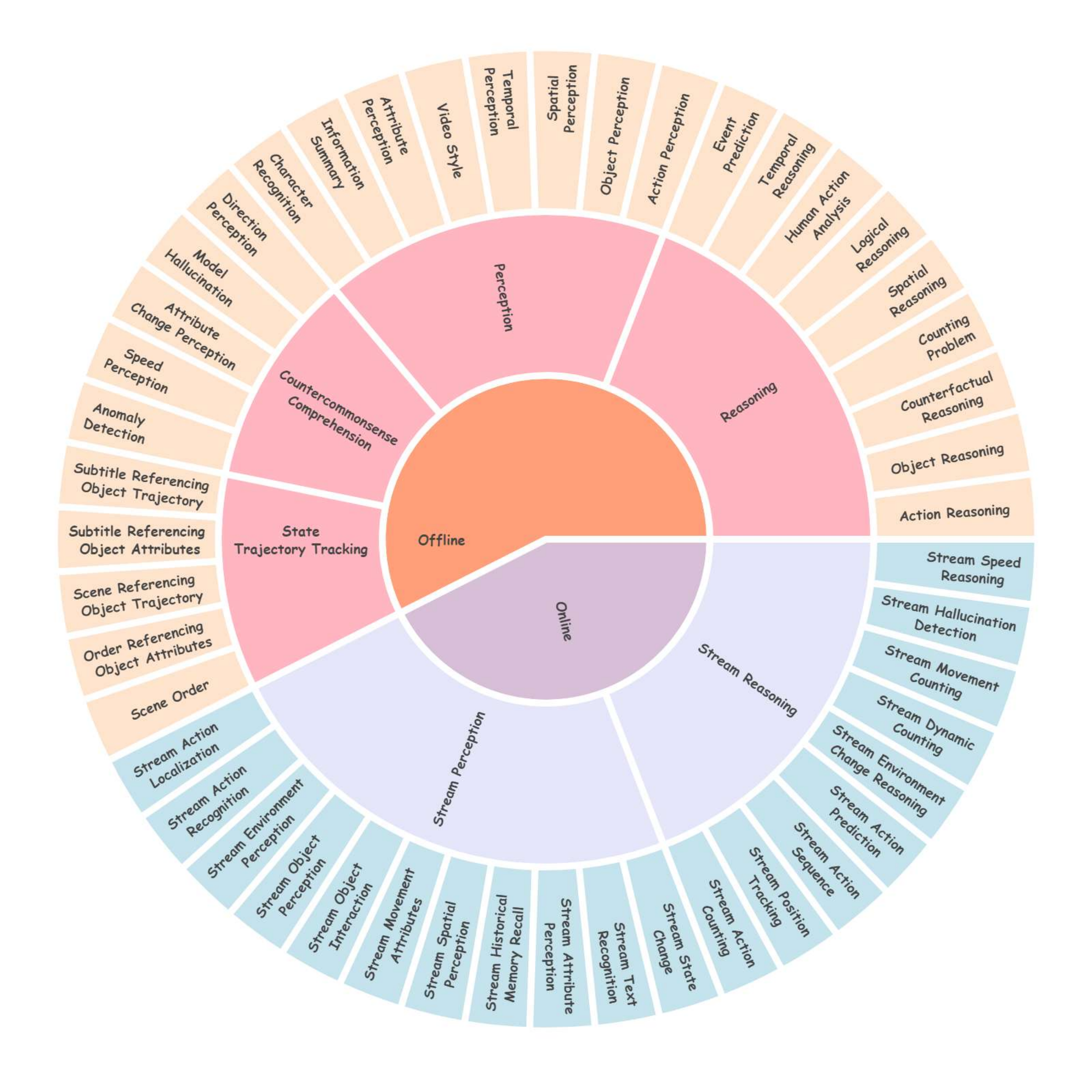} 
    \label{fig:short-a}
  \end{subfigure}
  \hfill
  \begin{subfigure}{0.38\linewidth}
    \includegraphics[width=\linewidth]{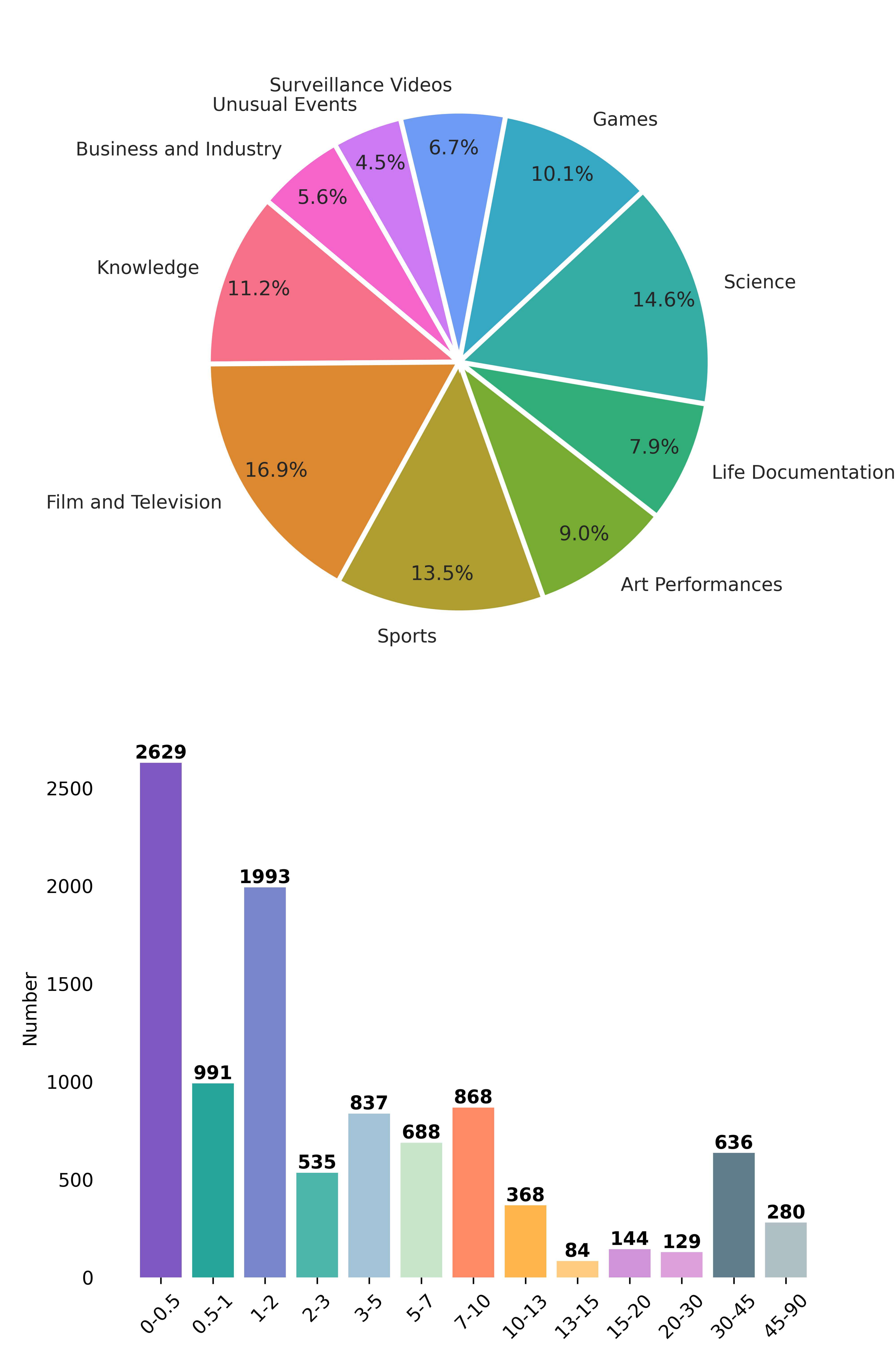} 
    \label{fig:short-b}
  \end{subfigure}
  \caption{(Left) Overview of ability dimensions in H²VU. Currently, H²VU incorporates three
levels of ability dimensions (L-1 to L-3), encompassing 47 distinct leaf abilities.
(Right) Video categories and Video duration length .H²VU covers 6 key domains, has a full spectrum of
video length and covers different core abilities of MLLMs.}
  \label{fig:short}
\end{figure*}
\subsection{Video Understanding Benchmarks.}
Driven by advancements in architecture, ongoing efforts focus on refining benchmark testing for Video \zql{understanding models} to guide the development of next-generation models. Previous research has integrated various evaluation aspects; traditional video benchmarks, such as MSVD-QA\cite{xu2017msrvtt_qa}, MSRVTT-QA\cite{xu2017msrvtt_qa}, and ActivityNet-QA\cite{yu2019activitynet}, primarily consist of short videos, typically lasting between one and two minutes. These datasets are meticulously annotated, including corresponding questions and correct answers. GPT-4\cite{openai2023gpt4} can be employed to assess answer accuracy by comparing model-generated responses with the questions and actual answers. However, these benchmarks mainly evaluate video understanding capabilities in static, short-scene contexts.

Consequently, researchers have proposed novel benchmarking methods for assessing logical reasoning and temporal understanding, exemplified by NExT-QA\cite{xiao2021next}, TemporalBench\cite{cai2024temporalbench}, and AutoEval-Video\cite{chen2023autoeval}. More recently, Video-MME\cite{fu2023mme}, LV Bench\cite{wang2024lvbench}, and LongVideo Bench\cite{wu2024longvideobench} have emerged, encompassing videos ranging from 20 minutes to over one hour, thus enabling the assessment of video understanding capabilities in long-video scenarios. In contrast, streaming video understanding requires models to sequentially process video streams and make decisions based on current and past information. This approach is particularly suited for scenarios where future data may be unavailable, such as embodied intelligence, autonomous driving, and augmented reality applications. Currently, Streaming Bench\cite{lin2024streamingbench} serves as the first comprehensive benchmark for evaluating multimodal large language model (MLLM) performance in understanding streaming video, revealing significant shortcomings in full-source and contextual understanding and providing critical directions for future research. OVO-Bench\cite{li2025ovo} emphasizes the importance of temporal awareness in online video understanding, categorizing it into three modes: proactive response, real-time visual perception, and retrospective analysis.

Our work builds upon the strengths of both traditional offline video benchmarks and general video task benchmarks to construct a new, high-quality video understanding benchmark, the H²VU-Benchmark. This benchmark aims to provide a more comprehensive perspective for evaluating video understanding models.benchmark aims to provide a more comprehensive perspective for evaluating video understanding models.

\section{H²VU-Benchmark}


\zql{In this section, we initially present an overview of the H²VU-Benchmark. Subsequently, we provide a detailed introduction to the benchmark collection, outlining the intended purposes of the evaluation tasks.  We then proceed with a thorough analysis of the statistics pertaining to the H²VU-Benchmark.}

\subsection{Overview}
\wq{To comprehensively assess the capabilities of video understanding models, the H²VU-Benchmark has developed a three-tier hierarchical competency classification system (L-1 to L-3), consisting of 10,183 evaluation tasks that encompass a broad spectrum of highly diverse data. 
The primary categories encompass two main areas: offline general video and online streaming video. \zql{For offline general videos, we employed common perception and reasoning tasks, alongside two additional tasks based on countercommonsense comprehension and trajectory tracking. For streaming videos, we utilized standard perception and reasoning tasks. Altogether, we have a total of 27 types of evaluation tasks for offline general videos and 20 types of evaluation tasks for online streaming videos.}
}

\subsection{Offline General Video \zql{Comprehension}}


\textbf{Perception.}\zql{As a fundamental component of video understanding, perception tasks are crucial for validating a model's ability to extract multi-level features from videos. We evaluate perception abilities across eight sub-dimensions from global to local levels. The sub-tasks that evaluate perception abilities on a global level are video style and information summarization, while the sub-tasks that evaluate perception abilities on a local level are action perception, object perception, spatial perception, temporal perception, attribute perception, and character recognition.}

\noindent{\textbf{Reasoning.}} \zql{Reasoning tasks aim to evaluate the model's ability to comprehend the underlying logic of a video. These tasks are assessed across nine detailed perspectives: action reasoning, object reasoning, counting,counterfactual reasoning spatial reasoning, logical reasoning, human activity analysis, temporal reasoning, and event prediction.}

\noindent\textbf{Countercommonsense \zql{Comprehension}.}
\zql{This task aims to thoroughly evaluate the model's comprehension abilities by focusing on its capacity for time-sequenced reasoning based on video content. To eliminate the influence of prior knowledge inherent in large language models, we have specifically designed tasks that violate general physical common sense and logic. These tasks include attribute change perception, speed perception, direction perception, model hallucination detection, and anomaly detection. Through these tasks, we intend to assess the model's proficiency in understanding and reasoning about dynamic scenarios presented in videos.}

\noindent\textbf{State Trajectory Tracking.}
\zql{This task centers on the continuous recognition and tracking of entities in videos. The challenge involves managing changes in appearance and shape, as well as periods of invisibility, necessitating the model to possess advanced perceptual capabilities. The task is composed of subtitle-based object trajectory tracking, subtitle-based object attribute changes, scene-based object trajectory tracking, scene-based object attribute changes and scene order reconstruction.}

\subsubsection{Online Streaming Video \zql{Comprehension}}
\zql{Streaming video presents a significant challenge due to the need for continuous updates, where questions may pertain to both recent segments and long-term information from the video. To address this, we have developed tasks that incorporate both real-time and historical perception and reasoning. These tasks are designed to comprehensively evaluate the model's understanding of streaming videos.}

\noindent\textbf{Online Perception.}
\zql{We propose a comprehensive evaluation framework for understanding streaming video perception by incorporating the following tasks: action localization, action recognition, environmental perception, object detection, object interaction, spatial perception, historical memory recall, attribute recognition, text detection, and state change identification. This multidimensional approach aims to assess video perception comprehension from various angles, ranging from detailed components to overall context, and from real-time analysis to memory-based understanding.}

\noindent\textbf{Online Reasoning.}
\zql{To evaluate the reasoning capabilities of streaming video, we propose a diverse set of tasks including action counting, location tracking, action sequencing, action prediction, environmental change inference, hallucination detection, and speed inference.}

\subsection{Dataset Construction}
\zql{We have amassed a substantial collection of videos. The offline general videos have primarily been sourced from existing datasets, whereas the streaming videos are partly derived from existing datasets with a significant portion being newly collected. To streamline the evaluation process, we converted these datasets into multiple-choice question-and-answer (QA) formats. These QA items were then categorized according to the hierarchical evaluation framework mentioned earlier.}

\begin{table*}[ht]
\centering
\begin{tabular}{llccccccccc}
\toprule
 \textbf{Benchmarks} &  & \textbf{\#Videos} & \textbf{\#Clips} & \textbf{\#ET} & \textbf{Len.(s)} & \textbf{\#QA Pairs} & \textbf{Anno.} & \textbf{Multi.} & \textbf{Open.} & \textbf{Online.}  \\
 \midrule
MSRVTT-QA \cite{xu2016msrvtt}& & 2,990 & 2,990 & 2 & 15.2 & 72,821  & A & \textcolor{red}{\xmark} & \textcolor{blue}{\checkmark} & \textcolor{red}{\xmark}  \\
MSVD-QA  \cite{xu2016msrvtt} & & 504   & 504   &2& 9.8  & 13,157  & A & \textcolor{red}{\xmark} & \textcolor{blue}{\checkmark} & \textcolor{red}{\xmark} \\
ActivityNet-QA \cite{yu2019activitynet} && 800   & 800 & 3  & 111.4 & 8,000   & M & \textcolor{red}{\xmark} &  \textcolor{red}{\xmark} & \textcolor{red}{\xmark} \\
\hline
MV Bench \cite{li2023mvbench} & & 3,641 & 3,641 & 20 & 16.0 & 4,000   & A & \textcolor{red}{\xmark}& \textcolor{blue}{\checkmark}  & \textcolor{red}{\xmark} \\
Video-Bench \cite{ning2023video} & & 5,917 & 5,917 & 3 & 56.0 & 17,036  & A\&M &\textcolor{red}{\xmark} & \textcolor{blue}{\checkmark} & \textcolor{red}{\xmark} \\
TempCompass \cite{Liu2024TempCompassDV}  & & 410   & 410 & 10 & 11.4 & 7,540   & A\&M & \textcolor{red}{\xmark} & \textcolor{red}{\xmark} & \textcolor{red}{\xmark} \\
Video-MME \cite{fu2023mme} & & 794   & 794 & 13& 18.4 & 2,700   & M & \textcolor{blue}{\checkmark} & \textcolor{blue}{\checkmark} & \textcolor{red}{\xmark} \\
VideoVista\cite{li2024videovista} & & 894   & 3,402 & 27& 131.0 & 24,906  & A & \textcolor{blue}{\checkmark} & \textcolor{blue}{\checkmark} & \textcolor{red}{\xmark} \\
\hline
Streaming Bench\cite{lin2024streamingbench}  & & 900 &- & 18 & 256& 4500 &  A\&M  &\textcolor{red}{\xmark}&\textcolor{blue}{\checkmark}&\textcolor{blue}{\checkmark}\\

OVO Bench\cite{li2025ovo}  & & 644 &- & 12 & 428&  2814 &  A\&M  &\textcolor{blue}{\checkmark}&\textcolor{blue}{\checkmark}&\textcolor{blue}{\checkmark}\\

\hline
\textbf{H²VU } & & 5902 & 5902 & 48 & 460.1 & 10183 & A\&M & \textcolor{blue}{\checkmark} & \textcolor{blue}{\checkmark} & \textcolor{blue}{\checkmark} \\
\bottomrule
\end{tabular}%
\caption{Comparisons across various benchmarks encompass several key aspects: 
Total number of videos (\#Videos), 
Number of clips (\#Clips), 
Number of evaluation tasks (\#ET), 
Average video length (Len.), 
Number of question-answer pairs (\#QA Pairs), 
Annotation method (Anno., with M/A representing Manual/Automatic methods), 
Whether videos cover multiple duration levels (Multi.), 
Whether videos originate from a broad open domain (Open.), 
Whether the benchmark tests online streaming video scenarios (Streaming).}
\label{tab:video_qa_benchmarks}
\end{table*}
\subsubsection{General Video Data \zql{Curation}}
\zql{To curate a high-quality video dataset, we have developed a three-step data curation methodology: static scene filtering, dialogue content recognition, and prior knowledge dependency purification. }

\zql{Static scene filtering utilizes optical flow motion analysis by utilizing the Farneback dense optical flow method\cite{farneback2003two} to compute motion fields between consecutive frames in a video sequence. A dynamic threshold, $\tau$ = 0.2, is established as the criterion for determining motion intensity. By calculating frame-by-frame optical flow vector magnitude statistics for candidate videos, we can identify and discard segments where the average optical flow value falls below $\tau$, thereby marking them as invalid static scenes. This process ensures the retention of dynamic videos, thereby maintaining the validity of the dynamic perception assessment.}

Dialogue content recognition leverages the spatiotemporal understanding capabilities of the Gemini 1.5 Flash\cite{team2023gemini}. Through self-regressive generative questioning, structured prompts (e.g., "Is the video primarily composed of dialogue between characters?") are input for three rounds of independent validation. If the dialogue content is classified as purely interview-based, it is labeled \zql{accordingly} and excluded. \zql{This process ensures that only narrative videos, which hold significant visual reasoning value, are retained.}

\zql{Knowledge dependency purification aims to eliminate the impact of large language models' prior knowledge on video understanding. To achieve this, we have developed a zero-shot experiment designed to filter out common-sense questions from textual question-answering benchmarks, thereby refining our focus on cross-modal understanding. Our approach involves inputting purely text-based questions into a powerful multimodal model, Gemini-1.5-Flash, for evaluation. If the model successfully answers a given question five consecutive times, we determine that video comprehension is not necessary for that particular question. Consequently, we remove the corresponding question-answer pair from the dataset.}

\subsubsection{Online Video Data \zql{Curation}}
\zql{The construction of this dataset involves three primary steps: video collection, question-answer annotation generation, and quality review. We manually collected 2,710 first-person videos using mobile phones. Key events within these videos were identified, and questions were formulated at the timestamps marking the end of these key events to simulate a streaming video Q\&A scenario. This process resulted in the creation of 4,000 question-answer pairs.}

\zql{To generate these question-answer pairs, we enlisted the expertise of four data annotation specialists, adhering to a structured methodology to produce open-ended questions and answers. Initially, the annotators watched the entire video to gain an in-depth understanding of its content. Next, they pinpointed visually salient regions through meticulous frame-by-frame analysis. Subsequently, open-ended question-answer pairs were crafted using a predefined task template.}

\zql{To enable a straightforward and quantitative evaluation of video comprehension, we utilized the GPT-4O Turbo model to convert open-ended answers into multiple-choice questions with four possible options. Specifically, along with the annotated answers, we employed a structured prompt strategy to generate three distractor choices. These distractors were designed based on the co-occurrence probabilities of visual entities, ensuring they were semantically related but contradictory to the content. The candidate questions thus generated were subjected to a review process conducted by four annotators to guarantee the production of high-quality question-answer pairs. The review focused on the questions' relevance to the video content, the logical coherence of the answer choices, and linguistic accuracy. Questions that were ambiguous or answerable without referencing the video content were eliminated to ensure that: (i) the language used was precise and clear; (ii) the questions were answerable, with realistic and logically consistent alternatives and correct answers.}

\zql{Additionally, to ensure the questions were sufficiently challenging and necessitated video content for accurate answers, we filtered out any question-answer pairs answerable solely based on text. These text-only questions were subsequently provided to the Gemini 1.5 Flash for further evaluation.
}

\subsection{Dataset Statistics}

\zql{The H²VU-Benchmark dataset comprises 5,902 video clips, which include 3,192 offline videos and 2,710 streaming videos, categorized into 10 various types. In total, there are 10,183 question-answer pairs associated with these videos. The duration of these videos ranges from as short as 1 second to as long as 90 minutes, with an average length of 406 seconds. The comprehensive distribution of the H²VU-Benchmark dataset across various task types, video categories, and durations is illustrated in Figure \ref{fig:short}. Next, we discuss the diversity of videos and evaluation tasks in our dataset with those in existing benchmarks, to highlight the comprehensiveness of our dataset evaluations.}

\zql{\noindent\textbf{Discussions.} As illustrated in Table \ref{tab:video_qa_benchmarks}, our benchmark demonstrates a significant advantage in terms of the number of videos, the number of clips, and the average video duration. This advantage is largely due to the diverse range of our data sources, which enables a more realistic and thorough evaluation of our benchmark. Additionally, our benchmark features high-quality question-and-answer pairs distributed across 47 tasks, derived through a combination of semi-automatic and manual selection processes. This approach has resulted in a substantial number of QA pairs, contributing to the reliability of our assessments. Furthermore, our dataset encompasses both streaming and offline videos, unlike other benchmarks that include only one or the other. This dual inclusion allows our benchmark to achieve a more comprehensive assessment.} 

\section{Experiments}

\begin{table*}[ht]
\centering
\renewcommand{\arraystretch}{1.2}
\newcommand{\ccell}[1]{\cellcolor{cyan!10} #1}
\newcommand{\mcell}[1]{\cellcolor{magenta!10} #1}
\newcommand{\ycell}[1]{\cellcolor{yellow!20} #1}
\newcommand{\gcell}[1]{\cellcolor{green!10} #1}
\newcolumntype{Y}{>{\raggedright\arraybackslash}X}

\begin{tabularx}{\textwidth}{lccYYYYYYYY}
\hline
Methods & Input  & Overall  & \multicolumn{5}{c}{Offline} & \multicolumn{3}{c}{Online} \\
\cmidrule(lr){4-8} \cmidrule(lr){9-11}
 & & & CC & STT & RE & PE & MEAN & ORE & OPE & MEAN  \\
\hline
\rowcolor{gray!10}
\textbf{\textit{Commercial MLLMs}}& &  &  &  &  &  &  &  &  &  \\
\ccell{GPT-4o \cite{openai2024gpt4o}} & \ccell{32 frm} & \ccell{67.0} & \ccell{61.1} & \ccell{46.8} & \ccell{67.3} & \ccell{69.1} & \ccell{65.8} & \ccell{70.2} & \ccell{69.1} & \ccell{69.6} \\
\ccell{Gemini-1.5-Flash  \cite{team2023gemini}} & \ccell{1 fps} & \ccell{68.0} & \ccell{61.5} & \ccell{46.3} & \ccell{68.4} & \ccell{77.1} & \ccell{67.5} & \ccell{68.8} & \ccell{70.1} & \ccell{69.2} \\
\ccell{Gemini-1.5-Pro \cite{team2023gemini}} & \ccell{1 fps} & \ccell{\textbf{72.5}} & \ccell{\textbf{64.3}} & \ccell{\textbf{53.1}} & \ccell{\textbf{74.1}} & \ccell{\textbf{79.2}} & \ccell{\textbf{71.9}} & \ccell{\textbf{74.3}} & \ccell{\textbf{73.1}} & \ccell{\textbf{73.6}} \\
\hline
\rowcolor{gray!10}
\textbf{\textit{Open Source MLLMs}}  & &  &  &  &  &  &  &  &  &  \\
\mcell{LLaVA-OneVision \cite{li2024llava}} & \mcell{64 frm}& \mcell{65.1} & \mcell{53.9} & \mcell{46.4} & \mcell{68.1} & \mcell{66.7} & \mcell{63.4} & \mcell{71.1} & \mcell{65.2} & \mcell{68.8}  \\
\mcell{VideoLLaMA 3 \cite{zhang2025videollama}} & \mcell{128 frm}& \mcell{66.0} & \mcell{57.3} & \mcell{42.7} & \mcell{68.7} & \mcell{68.3} & \mcell{65.4} & \mcell{67.6} & \mcell{67.4} & \mcell{67.5} \\
\mcell{Qwen2-VL \cite{wang2024qwen2}}  &\mcell{768 frm}& \mcell{66.4} & \mcell{60.2} & \mcell{47.4} & \mcell{70.2} & \mcell{72} & \mcell{67.1} & \mcell{67.7} & \mcell{62.2} & \mcell{65.1} \\
\mcell{LLaVA-Video \cite{fu2024video}} & \mcell{64 frm}& \mcell{67.9} & \mcell{56.2} & \mcell{49.6} & \mcell{73.7} & \mcell{71.4} & \mcell{67.9} & \mcell{70.7} & \mcell{64.1} & \mcell{68.0} \\
\mcell{VideoChat-Flash \cite{li2024videochat}}  &\mcell{512 frm}& \mcell{69.4} & \mcell{60.6} & \mcell{51.9} & \mcell{74.3} & \mcell{72.2} & \mcell{69.5} & \mcell{71.4} & \mcell{66.3} & \mcell{69.3} \\
\mcell{InternVL2.5 \cite{chen2024internvl2}}& \mcell{48 frm} & \mcell{69.8} & \mcell{61.0} & \mcell{\textbf{52.8}} & \mcell{72.6} & \mcell{70.6}  & \mcell{68.5}& \mcell{\textbf{73.8}} & \mcell{\textbf{69.7}} & \mcell{\textbf{72.7}} \\
\mcell{Qwen2.5-VL \cite{bai2025qwen2}} & \mcell{768 frm}& \mcell{\textbf{69.9}} & \mcell{\textbf{62.4}} & \mcell{51.6} & \mcell{\textbf{74.9}} & \mcell{\textbf{72.6}} & \mcell{\textbf{70.3}} & \mcell{70.7} & \mcell{66.7} & \mcell{69.1} \\
\hline

\end{tabularx}
\caption{The overall evaluation results on the H²VU dataset, including the number of input frames (Frames), the overall evaluation score(Overall), and the general H²VU tasks (CC: Countercommonsense Comprehension, STT: State Trajectory Tracking, RE: Reasoning Tasks, PE: Perception Tasks, ORE: Online Reasoning Tasks, OPE: Online Perception Tasks). MLLMs utilized two input strategies in the evaluation: uniform sampling (N frm), which uniformly samples N frames from the video, and frame rate sampling (N fps), which samples N frames per second. † denotes the use of low resolution.}
\label{tab:result}
\end{table*}

\subsection{Settings}
\zql{Based on the H²VU Benchmark, we conducted an extensive evaluation of various multimodal large language models for video comprehension. This assessment included both commercial and leading open-source 7B models. Specifically, we evaluated commercial models such as GPT-4o, Gemini-1.5-Pro, and Gemini-1.5-Flash, as well as open-source models including LLaVA-Video, LLaVA-Onevision, Internvl2.5, VideoLLaMA3, Qwen2.5VL, and Qwen2-VL. All models were assessed using their official implementations with default hyperparameters or available APIs, and the evaluation was performed in a zero-shot manner. We utilized accuracy as the evaluation metric, determined by comparing the model outputs with the ground truth, without involving any third-party models.}

\subsection{Main Results}

\textbf{Comparison of Commercial and Open-Source Models}
\zql{As demonstrated in Table \ref{tab:result}, Gemini-1.5-Pro surpasses all MLLMs with the highest overall score of 72.5,With overall scores of 67.0 and 68.0, GPT-4o and Gemini-1.5-Flash  trail the top-performing model by a margin.Qwen2.5-VL achieves the highest overall score of 69.9 in Open Source MLLMs. Closely following are InternVL2.5 and VideochatFlash, with scores of 69.8 and 69.4, respectively. These results highlight their exceptional general capabilities in both offline video understanding and streaming comprehension. The remaining models score as follows: LLaVA-Video at 67.9, Qwen2-VL at 66.4, VideoLLaMA3 at 66.0, and LLaVA-Onevision at 65.1. Despite Gemini-1.5 Pro retaining its preeminence in overall capabilities, several open-source MLLMs, such as Qwen2.5-VL, have outperformed specific Commercial models in evaluation. This achievement accentuates the rapid progress of the open-source community in MLLMs. It demonstrates their nascent potential for competing with commercial systems through optimized architectures. }

\zql{\textbf{Challenges in State Trajectory Tracking and  Countercommonsense Comprehension for MLLMs. }Further analysis reveals significant differences in the performance of various models on state trajectory tracking and countercommonsense comprehension tasks. Their performance is notably lower compared to the reasoning and perception tasks. In the countercommonsense comprehension task, Qwen2.5-VL achieves the highest score of 62.37, while VideoLLaMA3 scores the lowest score of 57.28. This indicates that although some models perform relatively well, the overall level remains low, reflecting the current limitations of MLLMs in surpassing prior knowledge constraints and identifying unreasonable causal relationships.
In the state trajectory  tracking task, scores are generally low across models, with InternVL2.5 achieving the highest score of 52.84, while VideoLLaMA3 only scores 42.68. This suggests that in complex dynamic scenes, existing models generally perform poorly in tracking target states and trajectories. In contrast, the models perform relatively well in reasoning and perception tasks. For instance, VideoLLaMA3 scores 68.7 and 68.29 in reasoning and perception tasks, respectively, significantly higher than its performance in the state trajectory  tracking task.
Currently available open-source multimodal large language models perform poorly in state trajectory  tracking and countercommonsense comprehension tasks in complex dynamic scenes, necessitating further improvement and enhancement. }

\zql{\textbf{Lack of Scenario-specific Training for Online Streaming Video. }Additionally, we analyze the performance differences of the models in offline (regular video) and online (streaming video) settings. In online streaming videos, InternVL2.5 performs best. However, Qwen2.5-VL, which performs well in offline videos, has mediocre performance in online scenarios, scoring only 69.1, lower than VideochatFlash's 69.3 and InternVL2.5's 72.7. This indicates that some models were not optimized for streaming video scenarios during training, leading to diminished performance. This highlights the significant challenges video understanding models face in actual streaming scenarios. }

\zql{The experimental results indicate the significance of incorporating a diverse range of evaluation tasks and datasets, underscoring the need for thorough assessment of multimodal models' video understanding abilities in real-world contexts. We aspire for our benchmark to assist in the comprehensive and integrated capability testing of large multimodal models and to inspire these models to enhance their focus on video knowledge comprehension, dynamic object tracking, and streaming video analysis.}

\subsection{Analysis}

\begin{figure}[t]
  \centering
   \includegraphics[width=1.05\linewidth]{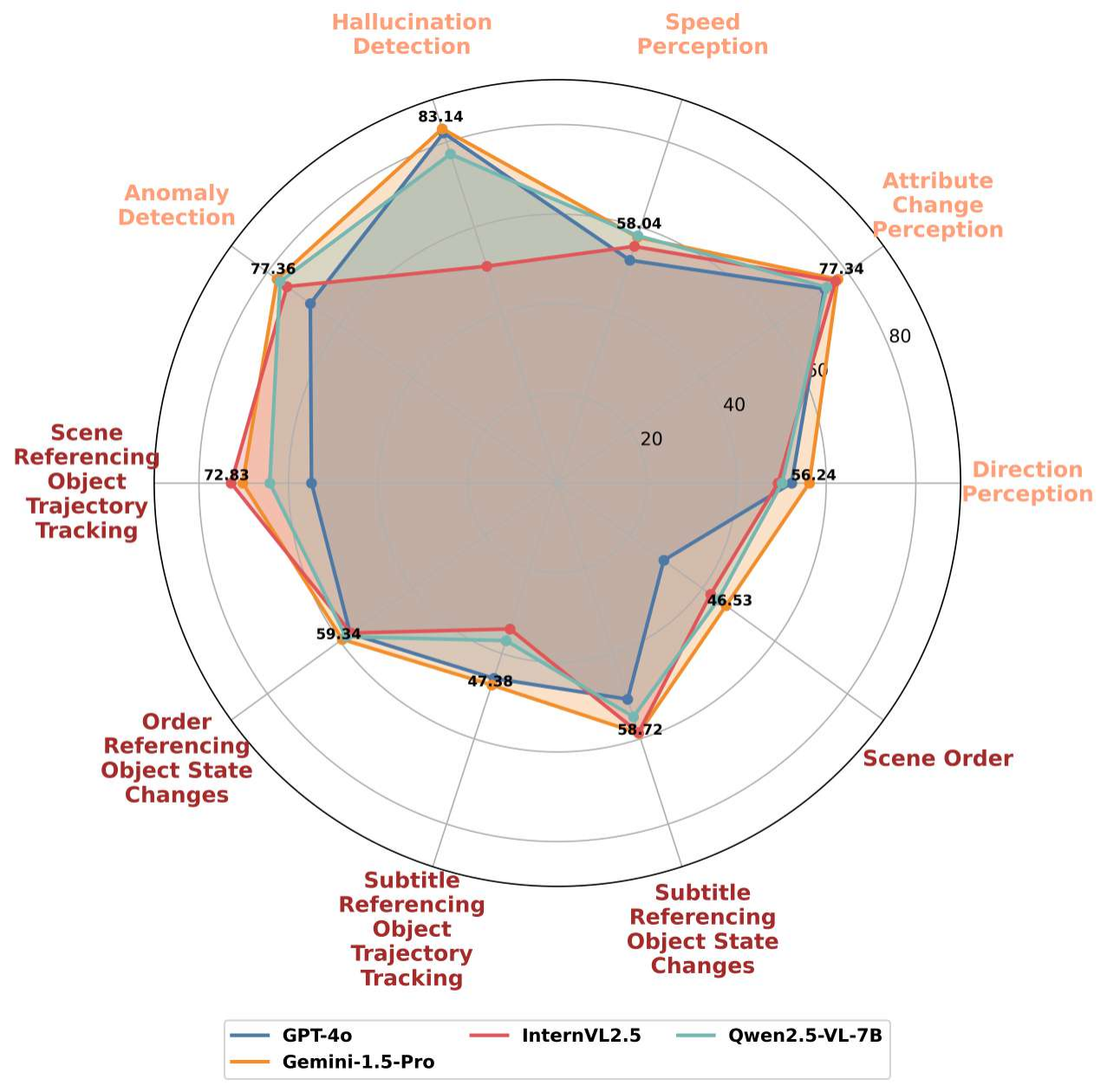}

   \caption{A comparison of commercial and open-source video large language models' performance on newly proposed tasks is shown. The figure details the average scores for each model on state trajectory tracking and countercommonsense comprehension.}
   \label{fig:comp}
\end{figure}

As shown in the fig \ref{fig:comp}, our in-depth investigation of Multimodal Large Language Models (MLLMs) across 10 subtasks within state trajectory tracking and countercommonsense comprehension tasks reveals critical performance disparities. 
First, commercial MLLMs exhibit significantly superior performance in hallucination detection compared to advanced open-source MLLMs (e.g., GPT-4o outperforms Qwen2.5-VL by more than 5 percentage points), suggesting targeted optimization strategies for hallucination mitigation in proprietary systems. 
This is also the optimization direction of current open-source MLLMs.

Moreover, GPT-4o, when restricted to $32$-frame inputs, underperforms leading open-source MLLMs in state trajectory tracking subtasks that require robust temporal understanding (e.g., GPT-4o lags behind other models by at least 10 percentage points in the task of scene order). \\
This deficiency is possibly attributable to the inherent limitations of sparse frame sampling, which inadequately captures fine-grained motion dynamics and inter-frame dependencies.

Furthermore, we conducted a qualitative evaluation of two cases in Figure \ref{fig:onecol} (SP: Speed Perception and SRS: Subtitle Referencing Object State Changes). These two cases comprehensively examined the model's capabilities in counter-intuitive OCR, attribute perception, action perception, object recognition, and temporal reasoning, thus being highly challenging.

In the case corresponding to SP, the video shows a stationary high-speed rail and a train rapidly departing from the platform. The question asked is "which train has a higher speed?". Gemini1.5 Pro correctly selected Option A, the train departing from the platform. However, GPT-4o, when input with 32 frames, chose the wrong option. When input with frames extracted, there may be an illusion that both the high-speed rail and the train are running in the pictures. In common life knowledge, the speed of high-speed rail is usually faster than that of a train, and the model's prior knowledge may further mislead to a wrong conclusion. This highlights the importance of reducing the dominant position of prior knowledge.

The event corresponding to SRT is the action states of two men in a fight. The question is "a man wearing a green coat throws a punch towards a man in black clothes. When the subtitle 'Accidentally,' appear, what change occurs to the man in black clothes?". The difficulty of this case lies in that the model first needs to have excellent object continuous perception ability to lock on the two fighting men. At the same time, it tests the collaborative operation of text OCR ability and action perception. Due to the short time when this subtitle appears, only Gemini1.5 Pro successfully noticed this frame and selected the correct option.

In conclusion, our task-specific evaluations expose MLLMs‘ fundamental reliances on prior knowledge and lack the capacity for continuous object perception and reasoning. Even advanced MLLMs demonstrate over-reliance on LLM priors for continuous object perception (e.g.,  erroneous trajectory predictions stemmed from ignoring observable motion cues) and directional velocity estimation ( failures correlated with text-based assumptions ).Therefore, we hold that Debiasing techniques to mitigate prior-knowledge dominances and enhancingthe MLLMs' persistent perception and reasoning ability are the one of  main development direction of MLLMs in the field of video understanding.

\section{Discussion \& Conclusion}

 H²VU has revealed several key insights into current Multimodal Large Language Models (MLLMs) and pointed out areas for future improvement. Here, we further discuss potential future development directions.

\textbf{Enhancing the Prior Knowledge Correction Capability and the Continuous understanding Ability of Multimodal Large Language Models.}
One significant challenge identified in our evaluation is that MLLMs perform significantly worse on tasks related to counter-commonsense understanding compared to other common tasks. In counter-commonsense situations, models need to rely less on prior knowledge and focus more on the content and temporal relationships within the video to reason correctly. This highlights the need for a corpus oriented towards counter-commonsense understanding, which can better utilize advanced architectural innovations to provide sufficient training supervision for multimodal large language models, thereby robustly reducing their reliance on prior knowledge. Similarly, all MLLMs show a certain gap in state trajectory tracking tasks compared to other common video understanding tasks. Therefore, whether from the perspective of data construction or architecture, innovative methods are needed to enhance the continuous perception and reasoning ability of multimodal large language models. 

\textbf{Targeted Optimization for Online Streaming Scenarios.}
In our evaluation, we found that some models, although leading in offline general videos, experience a performance drop when the scenario switches to online streaming videos. This may be because the models were not trained with data related to this scenario, which will pose certain obstacles to their application in the real world.

In this work, we \zql{present the} H²VU Bench\zql{mark}, a comprehensive benchmark that primarily addresses counter-commonsense understanding, state trajectory tracking tasks, and streaming video tasks, designed to evaluate the offline and online video comprehension capabilities of Video-LLMs. We \zql{believe} that H²VU will serve as a valuable resource for the research community, \zql{propelling} Video-LLMs toward practical, real-world applications. By \zql{identifying} current limitations and providing a rigorous evaluation platform, we \zql{aspire to} inspire future research dedicated to advancing video understanding and achieving human-level comprehension in AI systems.
{
    \small
    \bibliographystyle{ieeenat_fullname}
    \bibliography{main}
}

\end{document}